\documentclass{article}
\usepackage{titlesec}
\usepackage{cite}
\usepackage{enumitem}
\usepackage{authblk}
\usepackage{geometry}
\usepackage{subfigure}
\usepackage{multirow}
\usepackage{booktabs}
\usepackage{amsmath}
\usepackage{amssymb}
\usepackage{amsfonts}
\usepackage{stmaryrd}
\usepackage{algpseudocode}
\usepackage{array}
\usepackage{float}
\usepackage{textcomp}
\usepackage{subcaption}
\usepackage{ccicons}
\usepackage{arydshln}
\usepackage{hyperref}
\usepackage{graphicx}
\usepackage{doi}
\usepackage{listings}
\usepackage{tikz}
\usepackage{algorithm}
\usetikzlibrary{arrows.meta, positioning}

\title{\textbf{Opus} \vspace{0.5em} \\ \Large{\fontseries{bx} A Quantitative Framework for Workflow Evaluation}}

\author{
\vspace{1em}

\begin{minipage}[t]{0.25\textwidth}
    \centering
    \begin{tabular}[t]{c}
        \small{\textbf{Alan Seroul}}\\ \vspace{-0.7em}
        \scriptsize{Research Affiliate} \\
        \scriptsize{AppliedAI} \\
    \end{tabular}
\end{minipage}%
\begin{minipage}[t]{0.25\textwidth}
    \centering
    \begin{tabular}[t]{c}
        \small{\textbf{Théo Fagnoni}} \\ \vspace{-0.7em}
        \scriptsize{Member of Technical Staff} \\
        \scriptsize{AppliedAI} \\
    \end{tabular}
\end{minipage}%
\begin{minipage}[t]{0.25\textwidth}
    \centering
    \begin{tabular}[t]{c}
        \small{\textbf{Inès Adnani}} \\ \vspace{-0.7em}
        \scriptsize{Member of Technical Staff} \\
        \scriptsize{AppliedAI} \\
    \end{tabular}
\end{minipage}%
\vspace{\baselineskip}

\hspace*{0.125\textwidth}%
\begin{minipage}[t]{0.25\textwidth}
    \centering
    \begin{tabular}[t]{c}
        \small{\textbf{Dana O. Mohamed}}\\ \vspace{-0.7em}
        \scriptsize{Member of Technical Staff} \\
        \scriptsize{AppliedAI} \\
    \end{tabular}
\end{minipage}%
\begin{minipage}[t]{0.25\textwidth}
    \centering
    \begin{tabular}[t]{c}
        \small{\textbf{Phillip Kingston}$^*$}\\ \vspace{-0.7em}
        \scriptsize{Member of Technical Staff} \\
        \scriptsize{AppliedAI} \\
    \end{tabular}
\end{minipage}%
\hspace*{0.125\textwidth}
\vspace{1em}
}

\date{\normalsize{5 November 2025}}

\begin{document}

\maketitle
\begin{center}
    \ccbyncsa \\
    \vspace{0.3em}
    \footnotesize{This work is licensed under a Creative Commons Attribution-Noncommercial-ShareAlike 4.0 International License (CC BY-NC-SA 4.0)}
\end{center}

\footnotetext{$^*$ Corresponding author: phillip.kingston@opus.com}
\footnotetext{$^1$ Opus: \href{https://www.opus.com}{opus.com}}

\vspace{1em}

\begin{abstract}

\noindent 
This paper introduces the Opus Workflow Evaluation Framework, a probabilistic-normative formulation for quantifying Workflow quality and efficiency. It integrates notions of correctness, reliability, and cost into a coherent mathematical model that enables direct comparison, scoring, and optimization of Workflows. The framework combines the Opus Workflow Reward, a probabilistic function estimating expected performance through success likelihood, resource usage, and output gain, with the Opus Workflow Normative Penalties, a set of measurable functions capturing structural and informational quality across Cohesion, Coupling, Observability, and Information Hygiene. It supports automated Workflow assessment, ranking, and optimization within modern automation systems such as Opus$^1$ and can be integrated into Reinforcement Learning loops to guide Workflow discovery and refinement. In this paper:

\begin{enumerate}
\item We introduce the Opus Workflow Reward model that formalizes Workflow success as a probabilistic expectation over costs and outcomes.
\item We define measurable Opus Workflow Normative Penalties capturing structural, semantic, and signal-related properties of Workflows.
\item We propose a unified optimization formulation for identifying and ranking optimal Workflows under joint Reward–Penalty trade-offs.
\end{enumerate}
\end{abstract}

\section{Introduction}
Evaluating the quality and efficiency of work has been a core problem in automation since its inception. From early industrial systems to modern AI-driven pipelines, the need to evaluate how effectively tasks are executed, both by humans and machines, remains central to productivity, accountability, and optimization. Today, as organizations increasingly rely on Large Language Models (LLMs) and agents to plan, execute, and refine operational processes, the question of how to formally evaluate these processes becomes critical.

\vspace{1em}

Existing business frameworks, such as Business Process Management (BPM), process querying and mining, and related methods, provide tools for analyzing how work is structured and performed, yet they were not designed for probabilistic, AI-generated processes that exhibit uncertainty, continuous adaptation, and multi-agent interactions. NLP metrics such as BLEU, ROUGE, and BERTScore quantify textual overlap but are inadequate for assessing the structural and operational quality of deployed Workflows.

\vspace{1em}

To address this gap, we propose the Opus Workflow Evaluation Framework, which integrates probabilistic modeling with normative penalties to evaluate Workflows as stochastic, resource-bounded processes. The framework extends concepts from operations research, software engineering, and BPM into a quantitative formulation suitable for modern AI system evaluation at scale. It provides a principled basis for measuring, comparing, and optimizing Workflows according to their expected performance and structural quality, establishing Workflow evaluation as a measurable and computationally grounded discipline.

\paragraph{Definitions} Our methodology is based on the following concepts, a subset of which has been previously introduced in \textit{Opus: A Large Workflow Model for Complex Workflow Generation} by Fagnoni et al. \cite{fagnoni2024opus}, \textit{Opus: A Workflow Intention Framework for Complex Workflow Generation} by Kingston et al. \cite{kingston2025workflow}, and \textit{Opus: A Prompt Intention Framework for Complex Workflow Generation} by Fagnoni et al. \cite{fagnoni2025prompt}:

\vspace{1em}

\noindent\hangindent=2em\hangafter=0 \textbf{Workflow:} A Workflow is a Directed Acyclic Graph (DAG) of Task, Workflow Input, and Workflow Output nodes, with edges capturing data and execution dependencies. Workflows define ordered sequences of operations that transform inputs into target outputs. The same algorithm may be represented by multiple Workflows at varying levels of abstraction, granularity, or descriptive detail.

\vspace{1em}

\noindent\hangindent=2em\hangafter=0 \textbf{Workflow Input:}
Workflow Inputs are represented by dedicated input nodes in a Workflow. Each input node accepts a single, explicit data structure, ensuring clarity in how data is processed and transformed. Every input node is associated with a correctness probability, capturing the likelihood that the provided document conforms to the expected information and format. This accounts for cases where the input of one Workflow is produced by the output of another, and supports resilience against malformed data.

\vspace{1em}

\noindent\hangindent=2em\hangafter=0 \textbf{Workflow Output:}
Workflow Outputs are represented by dedicated output nodes in a Workflow. Each output node emits a single, explicit data structure, ensuring interpretability and a clear mapping between process and outcome. Every output node is associated with a correctness probability, derived from the overall information and dependencies within the Workflow.

\vspace{1em}

\noindent\hangindent=2em\hangafter=0 \textbf{Workflow Task:} An atomic unit of work within a Workflow, performing a specific function with defined input and output schemas, objectives, timing constraints, and success criteria. Tasks follow a single-responsibility principle (defined later in the paper), support automation or manual intervention, and maintain contextual awareness of dependencies. Tasks are auditable by humans or AI agents against their definition.

\vspace{1em}

\noindent\hangindent=2em\hangafter=0 \textbf{Task cumulative resources:} Resources permanently consumed by the execution of a Workflow Task, accumulating over the course of a Workflow execution. Examples include monetary cost, energy usage, persistent storage (e.g. hard drive space), raw materials, mobile data on capped plans, and labor hours.

\vspace{1em}

\noindent\hangindent=2em\hangafter=0 \textbf{Task releasable resources:} Resources temporarily allocated during the execution of a Workflow Task, held for its duration and released upon completion. Examples include RAM, CPU or GPU cycles, active threads, API quota, bandwidth, manpower, etc.

\vspace{1em}

\noindent\hangindent=2em\hangafter=0 \textbf{Workflow gain:}
The measurable value produced when a Workflow executes successfully. It represents the benefit, typically monetary, obtained by automating a process execution that would otherwise require longer execution time and higher per-hour costs. It may also capture benefits such as time savings, reduced risk, or improved accuracy, expressed on a consistent value scale.

\section{Background}
\paragraph{A historical challenge: measuring the quality of work}
The evaluation of work quality has long been a central concern in industrial and organizational systems. Following the industrial revolution, Taylor’s theory of scientific management sought to turn labor efficiency into a measurable science, focusing on standardization, time-motion analysis, and waste elimination. As economies evolved through mass production, digitalization, and globalization, the challenge of measuring and improving work quality spread across fields such as operations research and software engineering. Business Process Management (BPM) emerged as a discipline defining process models as structured representations of organizational Workflows and introducing analytical frameworks for their evaluation \cite{mendling_metrics_for_bpm, dumas_singular_responsibility, quality_in_bpm, cardoso_control_flow_cfc}. In parallel, process querying and mining techniques \cite{polyvyanny_process_query_language, polyvyanny_process_mining_logs, aalst_workflow_management} enabled the comparison of real executions with their modeled intent, bridging the gap between theory and practice. These developments established a foundation for assessing structure and correctness but did not provide means to quantify expected performance, uncertainty, or value, which are essential in modern AI-driven contexts.

\paragraph{Frameworks and derivations from software engineering}
In BPM literature, Workflows are viewed as structured models evaluated through principles of structural soundness and semantic clarity. The ISO 9000 framework defines quality as “the degree to which a set of inherent characteristics fulfills requirements”, aligning naturally with the evaluation of process models. From software engineering, principles such as the Single Responsibility Principle (SRP), Cohesion, and Coupling offer parallel insights: atomicity, modularity, and maintainability are as relevant to Workflow graphs as to code. Vanderfeesten et al. \cite{mendling_metrics_for_bpm} map software-quality metrics to process models, while Petri net properties such as soundness, liveness, and safeness \cite{aalst_workflow_management} provide tools for ensuring correctness. Together, these foundations establish a rigorous structural basis for Workflow evaluation.

\paragraph{Limitations of current approaches}
Despite these advances, classical BPM and software metrics largely assume deterministic execution. Metrics such as control-flow complexity \cite{cardoso_control_flow_cfc}, modularity indices \cite{mendling_metrics_for_bpm}, or SEQUAL-style quality dimensions \cite{quality_in_bpm} cannot express probabilistic dependencies, cumulative cost propagation, or stochastic reliability. Similarly, semantic evaluation methods (BLEU, ROUGE, METEOR, and BERTScore) assess textual similarity but not operational fidelity. These tools therefore fail to capture whether a deployed Workflow performs efficiently, reliably, or intelligibly in practice.

\paragraph{Bridging BPM and probabilistic evaluation}
Recent work in AI-assisted automation reframes Workflows as DAGs of stochastic Tasks, each with measurable probabilities of success, durations, and resource vectors. Evaluating such systems requires unifying the structural clarity of BPM with stochastic modeling. This intersection calls for a framework capable of quantifying expected performance under uncertainty while retaining interpretability and normative alignment.

\paragraph{The Opus quantitative approach}
The Opus Workflow Evaluation Framework addresses this challenge by formalizing Workflows as measurable stochastic systems. Each Task is modeled as a random variable over a measurable space, while Workflow performance and quality are expressed through expectation-based Reward and Normative Penalty functions. The Reward function quantifies the expected value of a Workflow given its success probabilities, resource costs, and gains; the Normative Penalties extend classical principles into continuous, measurable forms (Cohesion, Coupling, Observability, and Information Hygiene). Collectively, they establish a quantitative basis for comparing, ranking, and optimizing Workflows under uncertainty.

\section{Opus Workflow Evaluation Framework}
\subsection{Opus Workflow Formalism}
\paragraph{Workflow Task}
We model the execution of a Task as a random variable $X: \Omega \to \Omega'$, where $\Omega$ belongs to a probability space $(\Omega, \mathcal{A}, \mathbb{P})$ and $\Omega'$ is a measurable space. The sample space $\Omega$ represents the universe of all possible initial conditions for the Task. Note that both $\Omega$ and $\Omega'$ depend on the considered Task $X$. It is defined as the Cartesian product $\Omega = I \times S$, where: 

\vspace{1em}

$I$ is the Task input space. Each element $i \in I$ is a tuple where an input payload is paired with a boolean flag indicating its validity.

\vspace{1em}

$S$ is the set of all possible random seeds influencing the Task's stochastic behavior during its execution.

\vspace{1em}

An element $\omega = (i, s) \in \Omega$ thus represents a single, deterministic execution instance of the Task with a given input $i$ and a given seed $s$.

\vspace{1em}

The target space $\Omega'$ represents the space of all possible outcomes of the Task. It is defined as the Cartesian product $\Omega' = O \times R$, where:

\vspace{1em}

$O$ is the Task's output space. Each element $o \in O$ is a tuple of generated outputs, each accompanied by a boolean indicating its validity.

\vspace{1em}

$R \subseteq \mathbb{R}_+^{n+m+1}$ is a vector space representing the resources consumed during execution with $n, m \in \mathbb{N}$.

\vspace{1em}

The application of the Task $X$ to a specific initial condition $\omega = (i, s)$ yields a deterministic outcome $X(\omega) = (o, r)$, which consists of the output data $o$ and the consumed resources $r$. The probability measure $\mathbb{P}$ on $\Omega$ models the likelihood of encountering specific inputs and seeds, allowing us to reason about the probability distribution of outcomes in $\Omega'$. This probabilistic formalization is essential for analyzing a Task's success probability, performance, and overall quality.

\paragraph{Workflow}
We define a Workflow by a pair $(G, \Phi)$, where:

\vspace{1em}

$G = (V, E)$ is a DAG that encodes the dependency structure of the process. Nodes $v \in V$ represent either Workflow Inputs, Workflow Outputs or Workflow Tasks, and edges $e = (u \to v) \in E$ represent the execution flow and the data dependencies between them.

\vspace{1em}

The node set $V$ is partitioned into three disjoint subsets: $V = V_{\text{in}} \sqcup V_{\text{task}} \sqcup V_{\text{out}}$, where:
\begin{itemize}[label={}]
    \item $V_{\text{in}}$: Workflow Input nodes, which have no incoming edges. These nodes typically receive external data and serve as entry points into the Workflow.
    \item $V_{\text{out}}$: Workflow Output nodes, which have no outgoing edges. These nodes produce the final outcomes of the Workflow.
    \item $V_{\text{task}}$: Workflow Task nodes, which are internal and have both incoming and outgoing edges. These nodes perform intermediate transformations or decision making. 
\end{itemize}

\vspace{1em}

Each Workflow Input node $v \in V_{\text{in}}$ is associated with:
\begin{equation*}
\begin{aligned}
\phi_{\text{in}}:
\begin{cases}
V_{\text{in}} & \to [0,1] \times \mathcal{T} \\
v & \to \phi_{\text{in}}(v) = (\pi_v, \tau_v)
\end{cases}
\end{aligned}
\end{equation*}

where $\pi_v$ is the initial correctness probability, and $\tau_v \in \mathcal{T}$ denotes the type of data expected at node $v$. The type space $\mathcal{T}$ can be defined as a set of structured data schemas or type signatures (e.g. tuples, lists, or user-defined types with domain and codomain constraints).

\vspace{1em}

Each Workflow Task node $v \in V_{\text{task}}$ is associated with:
\begin{equation*}
\begin{aligned}
\phi_{\text{task}}:
\begin{cases}
V_{\text{task}} & \to [0,1] \times [0,1] \times \mathbb{R}^m_+ \times \mathbb{R}_+ \times \mathbb{R}^n_+ \times \mathcal{I}, \quad n, m \in \mathbb{N} \\
v & \to \phi_{\text{task}}(v) = (p_v, q_v, r^{(g)}_v, d_v, r^{(r)}_v, \iota_v)
\end{cases}
\end{aligned}
\end{equation*}
where:
\begin{itemize}[label={}]
    \item $p_v$: probability of success given that all parent Tasks succeed.
    \item $q_v$: probability of success given that at least one parent Task fails.
    \item $r^{(g)}_v \in \mathbb{R}^m_+$: cumulative resource vector.
    \item $d_v \in \mathbb{R}_+$: execution duration of the Task.
    \item $r^{(r)}_v \in \mathbb{R}^n_+$: releasable resource vector.
    \item $\iota_v \in \mathcal{I}$: Task implementation (a function), represented either as an effective probabilistic Turing machine or as its realization in a specific programming language. 
\end{itemize}

\vspace{1em}

We assume that the Workflows under comparison share an identical set of Workflow Inputs and Workflow Outputs. This ensures that differences in performance can be attributed solely to variations in their internal structure, without requiring additional domain-specific assumptions.

\vspace{1em}

We assume that for a given Task, its execution duration and resource consumption remain constant across its executions. This assumption yields a discrete timeline for Workflow execution, where each Task is either active or inactive, simplifying the evaluation of cumulative resource usage. In practice, however, this assumption often fails, e.g. in OCR Tasks where both duration and resource demand scale with document size. A better formulation can model duration and resources as functions of input characteristics. Nevertheless, data-driven estimation methods, based on historical execution traces and large-scale empirical measurements, can approximate these relationships with sufficient accuracy to preserve the validity of the framework under realistic conditions.

\subsection{Opus Workflow Reward}
\subsubsection{Workflow resource consumption}
Building on the Task-level resource definitions above, we measure Workflow-level resource consumption across three dimensions: permanently consumed cumulative resources that accumulate during execution, total execution time determined by the critical path of dependent Tasks, and releasable resources that must be provisioned concurrently at peak demand. This breakdown focuses on the main cost drivers of Workflow execution and provides a structured basis for evaluation and comparison.

\vspace{1em}

Let $V$ denote the node set of a Workflow $W = (G, \Phi)$, let $\tilde{\mathcal{P}}$ be the set of all root-to-leaf paths in $G$. We define the following Workflow-level resource consumption measures:

\begin{enumerate}
    \item Cumulative resources $R^{(g)}$: the total amount of permanently consumed resources throughout the Workflow,
    \begin{equation}
        R^{(g)}(W) = \sum_{v \in V_{\text{task}}} r^{(g)}_v
    \end{equation}
    \item Execution duration $d$: the worst-case (longest path) execution time, assuming sequential dependencies,
    \begin{equation}
        d(W) = \max_{P \in \tilde{\mathcal{P}}} \sum_{v \in P \cap V_{\text{task}}} d_v
    \end{equation}
    \item Releasable resources $R^{(r)}$: the peak concurrent demand of Workflow Task releasable resources at any time $t$, with $V_t$ denoting the set of possible active Tasks at $t$. We assume an ASAP schedule where each Task begins immediately after all predecessors have been completed, \begin{equation}
        R^{(r)}(W) = \max_{t} \sum_{v \in V_t \cap V_{\text{task}}} r^{(r)}_v
    \end{equation}
\end{enumerate}
More details are given in the Appendix.

\subsubsection{Success Probability}
Let $v$ be a node with parents $\{u_1,\dots,u_k\}$.  
For each node $v$, we define the event 
\begin{equation*}
    T_v := \{\text{``node $v$ produced a correct output"}\}
\end{equation*}

Let $p_v = \mathbb{P}(T_v \mid \bigcap_{j=1}^k T_{u_j})$ and  $q_v = \mathbb{P}(T_v \mid \bigcup_{j=1}^k \bar{T_{u_j}})$. 

\vspace{1em}

We assume the following:
\begin{enumerate}
    \item[(i)] Parent correctness events are independent: $ \mathbb{P}(\;\textstyle\bigcap_{j=1}^k T_{u_j}\;) = \prod_{j=1}^k \mathbb{P}(T_{u_j})
    $.
    \item[(ii)] Output nodes introduce no additional error modes. They act as logical conjunctions of their parents, which is equivalent to setting $p_v=1$ and $q_v=0$.
\end{enumerate}

\paragraph{Node success probability}

Workflow Input nodes: correctness is given by $\pi_v$,
\begin{equation*}
\mathbb{P}(T_v) = \pi_v
\end{equation*}

Workflow Task nodes: by the law of total probability and the independence assumption (i),
\begin{equation*}
\mathbb{P}(T_v) 
= q_v + (p_v - q_v)\prod_{j=1}^k \mathbb{P}(T_{u_j})
\end{equation*}
This models a two-phase behavior: if all parents are correct, the node succeeds with probability $p_v$; otherwise, it succeeds with a degraded probability $q_v$.

\vspace{1em}

Workflow Output nodes: correctness is the conjunction of parent correctness,
\begin{equation*}
\mathbb{P}(T_v) = \prod_{j=1}^k \mathbb{P}(T_{u_j})
\end{equation*}

\paragraph{Workflow success probability}
We define the Workflow success event as the simultaneous correctness of all Workflow Output nodes:
\begin{equation*}
T_W := \textstyle\bigcap_{v \in V_{\text{out}}} T_v
\end{equation*}
Accordingly, the Workflow success probability is
\begin{equation*}
\mathbb{P}(W) := \mathbb{P}(T_W) = \mathbb{P}(\; \textstyle\bigcap_{v \in V_{\text{out}}} T_v \;)
\end{equation*}
Exact evaluation requires the joint distribution of output correctness events. Assuming independence among $\{T_v, \; v \in V_{\text{out}}\}$, this reduces to
\begin{equation}
  \mathbb{P}(W) = \prod_{v \in V_{\text{out}}} \mathbb{P}(T_v)
\end{equation}

$\mathbb{P}(W)$ quantifies the likelihood that the Workflow produces entirely correct outputs given its inputs.

\subsubsection{Reward}
We define a scalar cost function that aggregates cumulative, temporal, and releasable resources:
\begin{equation}
  C(W) = \langle w^{(g)}, R^{(g)}(W) \rangle \;+\; w^{(d)}\,.\,d(W) \;+\; \langle w^{(r)}, R^{(r)}(W) \rangle
\end{equation}
where $w^{(g)} \in \mathbb{R}^m$, $w^{(d)} \in \mathbb{R}$, and $w^{(r)} \in \mathbb{R}^n$ are weights encoding the relative importance of each resource type.

\vspace{1em}

For each Workflow Output node $v \in V_{\text{out}}$, we define the random variable
\begin{equation*}
G_v = 
\begin{cases}
g_v & \text{with probability } \mathbb{P}(T_v) \\
0   & \text{with probability } 1 - \mathbb{P}(T_v)
\end{cases}
\end{equation*}
representing the gain realized if a Workflow Output is correct.

\vspace{1em}

The net benefit of executing the Workflow $W$ is then
\begin{equation}
  B_W = \sum_{v \in V_{\text{out}}} G_v \;-\; C(W)
\end{equation}
Fixed costs $C(W)$ are always incurred, while benefits $g_v$ are realized only when the corresponding Workflow Output succeed. We assume that the total
$C(W)$ is fixed for each Workflow execution, accounting for all Tasks defined in the Workflow, even those that may not run if execution halts prematurely. A more refined formulation could model per-Task costs as random variables, incurring cost only for Tasks that are actually executed.

\vspace{1em}

We define the expected Reward of $W$:
\begin{align}
  \mathcal{R}(W) &= \mathbb{E}[B_W] \nonumber \\
  &= \sum_{v \in V_{\text{out}}} \mathbb{P}(T_v)\,.\, g_v \;-\; C(W)
\end{align}

$\mathcal{R}(W)$ quantifies the expected net value of a Workflow execution, combining output-specific gains weighted by their success probabilities and offset by execution costs. Positive Reward indicates that the Workflow is expected to generate value; negative Reward indicates expected loss. The weights $(w^{(g)}, w^{(d)}, w^{(r)})$ can be set from empirical measurements, domain-specific heuristics, or market-based resource pricing, enabling comparisons across heterogeneous Workflows in operationally meaningful units.

\newpage

\subsection{Opus Workflow Normative Penalties}

We evaluate nodes and Workflows along four dimensions, referred to as the Opus Workflow Normative Penalties. Cohesion (Ch) penalizes nodes that conflate multiple responsibilities, reflecting the dispersion of functions within a node. Coupling (Cp) penalizes dependencies among nodes, capturing structural and semantic links that can propagate failures or complicate evolution. Observability (Ob) penalizes opacity and incompleteness in runtime signals, measuring the difficulty of inferring the actual state of execution from logs, metrics, and traces. Information Hygiene (Ih) penalizes the emission of irrelevant, redundant, or privacy-sensitive signals, ensuring that the information produced is concise, relevant, and actionable.

\vspace{1em}

From these dimensions we derive two higher-level evaluative penalties, which we postulate as normative requirements for Workflow quality. The Cohesive Independence Penalty (CIP) combines Cohesion and Coupling, ensuring that each node is atomic in purpose and minimally dependent on others. The Signal Integrity Penalty (SIP) combines Observability and Information Hygiene, ensuring that runtime signals are trustworthy, relevant, necessary, and sufficient for monitoring, debugging, and auditing. Both penalties are measurable and continuous: they provide criteria by which Workflows can be systematically compared and optimized, with 0 representing the optimal penalty-free state and 1 representing the worst-case scenario.

\vspace{1em}

Let $W \in \mathcal{W}$ denote a Workflow with $|V_{\text{task}}| = n$ Workflow Task nodes.

\paragraph{Cohesion Penalty (Ch)}
Cohesion denotes the degree to which each node in a Workflow performs a single, well-defined function. A cohesive node is singular and atomic; its function can be specified without ambiguity or the mixing of unrelated concerns.

\vspace{0.5em}

A low Cohesion Penalty provides clarity of purpose and supports maintainability. A high Cohesion Penalty results in overloaded nodes that conflate responsibilities, creating ambiguity and complicating evolution.

\begin{equation*}
\begin{aligned}
\text{For a Task, Ch}:\;
\begin{cases}
V_{\text{task}} & \to [0,1] \\
v & \mapsto \text{Ch}(v)
\end{cases},
&\quad\quad
\text{for a Workflow, Ch}:\;
\begin{cases}
\mathcal{W} & \to [0,1] \\
W & \mapsto \big( \frac{1}{n} \sum_{v \in V_{\text{task}}} \text{Ch}(v)^2 \big)^{1/2}
\end{cases}
\end{aligned}
\end{equation*}

\paragraph{Coupling Penalty (Cp)}
Coupling measures the degree of dependence among nodes. It reflects both structural connections and semantic links such as shared data, state, or control.

\vspace{0.5em}

A low Coupling Penalty supports modularity and substitution: nodes can be altered, replaced, or reused without destabilizing the Workflow as a whole. A high Coupling Penalty produces brittle systems, where failures or modifications propagate widely.

\begin{equation*}
\begin{aligned}
\text{For a Task, Cp}:\;
\begin{cases}
V_{\text{task}} & \to [0,1] \\
v & \mapsto \text{Cp}(v)
\end{cases},
&\quad\quad
\text{for a Workflow, Cp}:\;
\begin{cases}
\mathcal{W} & \to [0,1] \\
W & \mapsto \big( \frac{1}{n} \sum_{v \in V_{\text{task}}} \text{Cp}(v)^2 \big)^{1/2}
\end{cases}
\end{aligned}
\end{equation*}

\newpage

\paragraph{Observability Penalty (Ob)}
Observability measures whether runtime signals (outputs, logs, traces, metrics) are sufficient and whether they faithfully reflect the Workflow’s actual execution state. It characterizes the alignment between exposed information and underlying execution.

\vspace{0.5em}

A low Observability Penalty indicates that signals are complete and accurate enough to support monitoring, diagnosis, and decision-making. A high Observability Penalty indicates that information is missing to establish the true execution state.

\begin{equation*}
\begin{aligned}
\text{For a Task, Ob}:\;
\begin{cases}
V_{\text{task}} & \to [0,1] \\
v & \mapsto \text{Ob}(v)
\end{cases},
&\quad\quad
\text{for a Workflow, Ob}:\;
\begin{cases}
\mathcal{W} & \to [0,1] \\
W & \mapsto \big( \frac{1}{n} \sum_{v \in V_{\text{task}}} \text{Ob}(v)^2 \big)^{1/2}
\end{cases}
\end{aligned}
\end{equation*}

\paragraph{Information Hygiene Penalty (Ih)}
Information Hygiene evaluates whether runtime signals are necessary and restricted to what is relevant for understanding the Workflow’s execution. It penalizes the emission of irrelevant, redundant, or privacy-sensitive information.

\vspace{0.5em}

A low Information Hygiene Penalty means that signals are concise, meaningful, and privacy-preserving. A high Information Hygiene Penalty means that signals are excessive, noisy or low-value, burdening operators and automated systems while obscuring critical insights.

\begin{equation*}
\begin{aligned}
\text{For a Task, Ih}:\;
\begin{cases}
V_{\text{task}} & \to [0,1] \\
v & \mapsto \text{Ih}(v)
\end{cases},
&\quad\quad
\text{for a Workflow, Ih}:\;
\begin{cases}
\mathcal{W} & \to [0,1] \\
W & \mapsto \big( \frac{1}{n} \sum_{v \in V_{\text{task}}} \text{Ih}(v)^2 \big)^{1/2}
\end{cases}
\end{aligned}
\end{equation*}

Note that Ch($v$), Cp($v$), Ob($v$) and Ih($v$) implicitly depend on the Workflow.

\paragraph{Cohesive Independence Penalty (CIP)}  
CIP integrates Cohesion and Coupling Penalties.
A Workflow achieves low CIP when its nodes each hold a clear, atomic responsibility (low Ch) and remain minimally dependent on others (low Cp).
CIP establishes structural modularity and functional separation as essential conditions of Workflow design.  

\vspace{1em}
Cohesion and Coupling are treated as complementary quantities of structural responsibility. To enable further factorization, we suppose $\forall \; v \in V_{\text{task}}, \; \mathrm{Ch}(v) + \mathrm{Cp}(v) = 1$.

\begin{equation*}
\text{CIP}:\;
\begin{cases}
\mathcal{W} & \to [0,1] \\
W & \mapsto \big( \alpha_{\mathrm{Ch}} \cdot \mathrm{Ch}(W)^2 + \alpha_{\mathrm{Cp}} \cdot \mathrm{Cp}(W)^2 \big)^{1/2}, \\
& \quad \alpha_{\mathrm{Ch}}, \alpha_{\mathrm{Cp}} \in \mathbb{R}_{+}, \; \alpha_{\mathrm{Ch}} + \alpha_{\mathrm{Cp}} = 1
\end{cases}
\end{equation*}

\vspace{1em}

$\forall \; W \in \mathcal{W},$  

\begin{equation}
\text{CIP}(W)^2 = \alpha_{\text{Ch}} \cdot \big(1-\alpha_{\text{Ch}}\big) + \frac{1}{n} \sum_{v \in V_{\text{task}}} \big(\alpha_{\text{Ch}} - \mathrm{Cp}(v)\big)^2
\label{eq:CIP}
\end{equation}

\newpage

\paragraph{Signal Integrity Penalty (SIP)}  
SIP integrates Observability and Information Hygiene Penalties.
A Workflow achieves low SIP when its runtime signals are both sufficient, accurate (low Ob) and necessary (low Ih) to reflect the underlying execution state.
SIP establishes transparency and reliability as essential conditions for trustworthy Workflow execution.

\vspace{1em}

Observability and Information Hygiene are treated as complementary quantities of signal integrity. To enable further factorization, we suppose $ \; \forall \; v \in V_{\text{task}}, \; \mathrm{Ob}(v) + \mathrm{Ih}(v) = 1$.

\begin{equation*}
\text{SIP}:\;
\begin{cases}
\mathcal{W} & \to [0,1] \\
W & \mapsto \big( \alpha_{\mathrm{Ob}} \cdot \mathrm{Ob}(W)^2 + \alpha_{\mathrm{Ih}} \cdot \mathrm{Ih}(W)^2 \big)^{1/2}, \\
& \quad \alpha_{\mathrm{Ob}}, \alpha_{\mathrm{Ih}} \in \mathbb{R}_{+}, \; \alpha_{\mathrm{Ob}} + \alpha_{\mathrm{Ih}} = 1
\end{cases}
\end{equation*}

\vspace{1em}

\begin{equation}
\text{SIP}(W)^2 = \alpha_{\text{Ob}} \cdot \big(1-\alpha_{\text{Ob}}\big) + \frac{1}{n} \sum_{v \in V_{\text{task}}} \big(\alpha_{\text{Ob}} - \mathrm{Ih}(v)\big)^2 
\end{equation}

\vspace{1em}

CIP and SIP together provide a measurable and normative foundation for Workflow evaluation.  
They define continuous criteria for assessing quality, enabling systematic comparison and principled optimization.

\paragraph{Opus Workflow Penalty}
We introduce

\begin{equation*}
\forall \; W \in \mathcal{W}, \quad \mathcal{L}(W)^2 = \gamma_s \cdot \text{CIP}(W)^2 + \gamma_d \cdot \text{SIP}(W)^2
\end{equation*}

where $\gamma_s, \gamma_d \in [0,1]$ are user-defined trade-off parameters.  

\vspace{1em}

Posing $\gamma_s + \gamma_d = 1$ without loss of generality, we have:

\begin{equation}
\mathcal{L}(W)^2 = \gamma_s \cdot \text{CIP}(W)^2 + (1-\gamma_s) \cdot \text{SIP}(W)^2
\end{equation}

By construction, $\mathcal{L}(W)$ is bounded between $0$ and $1$:
\begin{itemize}[label={}]
    \item $\mathcal{L}(W) = 0$ if the Workflow is perfectly minimized against all Penalty terms,
    \item $\mathcal{L}(W) = 1$ if the Workflow deviates maximally in all dimensions.
\end{itemize}

\begin{figure}[H]
\centering
\includegraphics[width=0.85\textwidth]{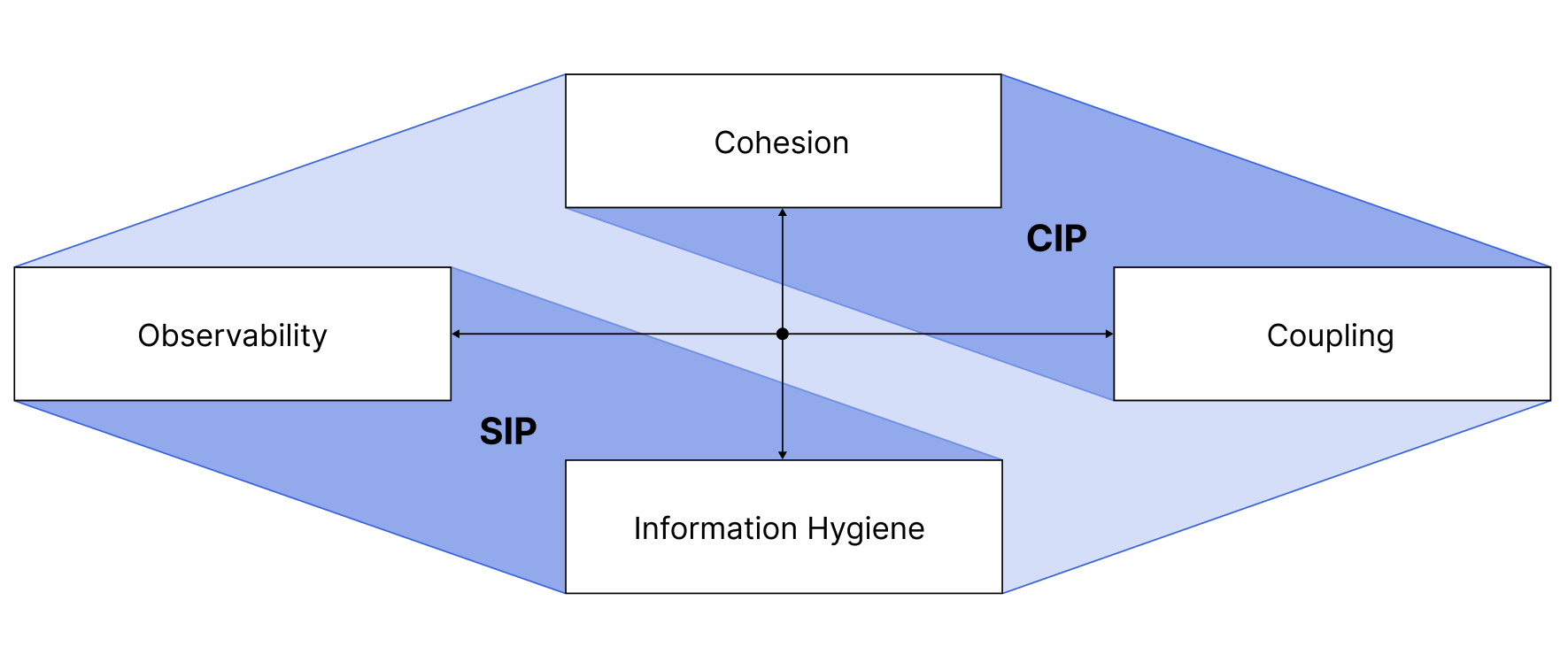}
\caption{\label{schema_properties} Opus Workflow Normative Penalties}
\end{figure}

These relationships are visualized in Figure ~\ref{schema_properties}, which represents how the four normative penalties (Ch, Cp, Ob, Ih) combine into the higher-level CIP and SIP penalties.

\subsection{Framework application}
\paragraph{Finding optimal Workflows}
One strategy for identifying optimal Workflows is a two-stage procedure: first maximize the Reward $\mathcal{R}$ to obtain efficient candidates, then, among these, minimize the Penalty $\mathcal{L}$ to select the best formulations.

\vspace{1em}

Formally, this process is expressed as the two-step optimization problem:
\begin{equation} \label{optim}
   \mathcal{W}^{**} =  \operatorname*{argmin}\limits_{W \in \mathcal{W}^*} \mathcal{L}(W)
\end{equation}
where
\begin{equation}
    \mathcal{W}^* = \operatorname*{argmax}\limits_{W \in \mathcal{W}_c} \mathcal{R}(W)
\end{equation}

$\mathcal{W}_c$ denotes the subset of feasible Workflows satisfying business and technical constraints (e.g. cost limits, manpower, bandwidth, reliability, and input/output specifications). These constraints may also include user-defined restrictions on performance or resources. Equation \eqref{optim} highlights the trade-off between maximizing Reward (raw performance) and minimizing Penalty (interpretability, complexity, maintainability, and structural quality). In this setting, the Reward reflects how well the current Workflow performs, while the Penalty measures its potential for further improvement.

\vspace{1em}

As with most optimization problems, the existence of a global optimum cannot be guaranteed, nor can any general algorithm ensure its discovery. Moreover, explicitly characterizing $\mathcal{W}_c$ is challenging. In practice, a common approach is to construct a finite candidate set $\mathcal{W}_c' \subset \mathcal{W}_c$ expected to perform well, then iteratively refine this set using heuristics, domain knowledge, or algorithmic exploration techniques.

\newpage

\paragraph{Ranking a set of Workflows}
While the framework permits comparisons between arbitrary Workflows, we recommend restricting comparisons to those sharing the same Workflow Inputs and Outputs and addressing the same business objective. This constraint ensures that compared Workflows operate within an equivalent process scope, making their evaluation and ranking both meaningful and practically relevant. Therefore, it should be enforced when constructing the candidate set $\mathcal{W}_c$.

\vspace{1em}

The reliability of ranking depends critically on consistent estimation of underlying parameters. For the comparison to be valid, resource costs and success probabilities must be assigned consistently across identical nodes in different Workflows. Otherwise, rankings may reflect parameter inconsistencies rather than genuine structural differences.

\vspace{1em}

Within a consistently estimated candidate set, Workflows can be ordered by a strict lexicographic preference: given two Workflows $W_1$ and $W_2$, we write
\begin{equation}
W_1 \succ W_2 \iff
\mathcal{R}(W_1) > \mathcal{R}(W_2)
\quad \text{or} \quad
\big(\mathcal{R}(W_1) = \mathcal{R}(W_2) \; \text{and} \;\mathcal{L}(W_1) < \mathcal{L}(W_2)\big)
\end{equation}

This rule reflects our optimization philosophy: Reward prevails, with Penalty acting as a tie-breaker.

\vspace{1em}

Beyond ordering, it is also useful to quantify performance gaps. We define the distance between two Workflows $W_1$ and $W_2$ as
\begin{equation}
d(W_1, W_2) = |\mathcal{R}(W_1) - \mathcal{R}(W_2)|
\end{equation}
which measures the absolute difference in Reward, consistent with the lexicographic order where Penalty is only considered when Rewards are equal.

\paragraph{A Reinforcement Learning Perspective}

The terms Reward and Penalty are deliberately borrowed from the Reinforcement Learning (RL) paradigm. This analogy highlights the inherently iterative nature of Workflow optimization, which can be described as a feedback-driven cycle:

\begin{enumerate}
    \item A candidate Workflow is proposed for a specific process.
    \item The candidate is assessed by estimating its resource costs, success probabilities, and expected gains.
    \item A Reward is computed, serving as a feedback signal that quantifies performance and encourages the discovery of superior alternatives.
    \item A Penalty is computed, guiding exploration toward well-structured Workflows that improve debugging, maintainability, and iterative refinement.
\end{enumerate}

In this perspective, the organization or system designing Workflows acts as the Agent, while the Opus Workflow Evaluation Framework provides the environmental signals through which outcomes are measured. The State is defined by the Reward and Penalty values associated with a candidate Workflow, the Action corresponds to proposing a new Workflow, and the Policy denotes the strategy or generative mechanism guiding the creation of such candidates.

\section{Case study}
To illustrate the framework in practice, we present a simple use case. This example benchmarks a set of candidate Workflows designed to automatically classify customer complaint emails.

\paragraph{Workflow Inputs}
The process requires the following inputs:
\begin{itemize}[label={}]
    \item \textit{Customer Full Name} (text): the customer’s complete name as registered in their profile, including both given and family names.
    \item \textit{Customer Email File} (email file): the raw email received from the customer, including headers and message body.
    \item \textit{Customer Account Number} (text): a unique customer account identifier, validated for existence and activity.
\end{itemize}

\paragraph{Workflow Output}
The expected output from the process is:
\begin{itemize}[label={}]
    \item \textit{Support Ticket Record} (support ticket JSON file): a structured ticket containing all required customer details and request information.
\end{itemize}

\paragraph{Data Characteristics}
We assume the following values for data and processing:
\begin{itemize}[label={}]
    \item Average email length: 100 tokens.
    \item Average prompt length: 100 tokens.
    \item Average response length: 50 tokens.
    \item Average LLM call duration: 1000 ms.
    \item Average Python script execution duration: 100 ms.
    \item Cold start latency (serverless warm-up): 800 ms.
    \item LLM costs:
    \begin{itemize}[label={}]
        \item Small model: input \$0.40 / 1M tokens (\$$4.0 \times 10^{-5}$ for 100 tokens), output \$1.60 / 1M tokens (\$$8.0 \times 10^{-5}$ for 50 tokens).
        \item Large model: input \$2.00 / 1M tokens (\$$2.0 \times 10^{-4}$ for 100 tokens), output \$8.00 / 1M tokens (\$$4.0 \times 10^{-4}$ for 50 tokens).
    \end{itemize}
\end{itemize}

\paragraph{Customer Lifetime Value (CLV)}
The business value depends on the customer segment:
\begin{itemize}[label={}]
    \item Small customer: \$30
    \item Medium customer: \$360
    \item Enterprise customer: \$16,000
\end{itemize}

\paragraph{Impact of Ticket Classification}
The quality of classification affects customer satisfaction:
\begin{itemize}[label={}]
    \item Well classified: the customer receives a relevant response quickly.
    \item Misclassified: three possible outcomes occur,
    \begin{itemize}
        \item No answer (5\% probability, 30\% churn).
        \item Delayed answer due to email redirection (80\% probability, 1\% churn).
        \item Uninformed answer (15\% probability, 5\% churn).
    \end{itemize}
\end{itemize}

\paragraph{Expected Loss Estimation}
The expected financial loss is expressed as:
\begin{align*}
g &= \text{CLV} \times (0.05 \times 0.3 + 0.8 \times 0.01 + 0.15 \times 0.05) \\
  &= \text{CLV} \times 0.0305
\end{align*}
Thus:
\begin{align*}
g_s &= 30 \times 0.0305 \approx \$0.92 \quad \text{(for small customer)}&\\
g_m &= 360 \times 0.0305 \approx \$10.98 \quad \text{(for medium customer)}&\\
g_e &= 16,000 \times  0.0305 \approx \$488 \quad\text{(for enterprise customer)}
\end{align*}

We interpret these quantities as the monetary value of avoiding the expected loss. We focus here on the B2C setting with a small customer (\$30 CLV).

\paragraph{Weights}
We assign:
\begin{align*}
    w^{(g)} &= 1 \quad (\text{all costs expressed in dollars}) \\
    w^{(d)} &= 2.1 \times 10^{-12} \quad (\text{serverless function execution cost per ms})
\end{align*}

\paragraph{Default Parameter Values}
Unless otherwise specified, the default values for $p$, $q$ and cost are $p = 1$, $q = 0$ and cost = 0. This simplified setting corresponds to a purely deterministic node with no stochastic uncertainty. For instance, retrieving an element from a dictionary always succeeds if the key exists and always fails otherwise. Cost is cumulative and expressed in dollars (\$). Duration (d) is expressed in milliseconds (ms).

\newpage

\subsection{Workflow 1}
\begin{figure}[H] 
    \centering
    \includegraphics[width=1\textwidth]{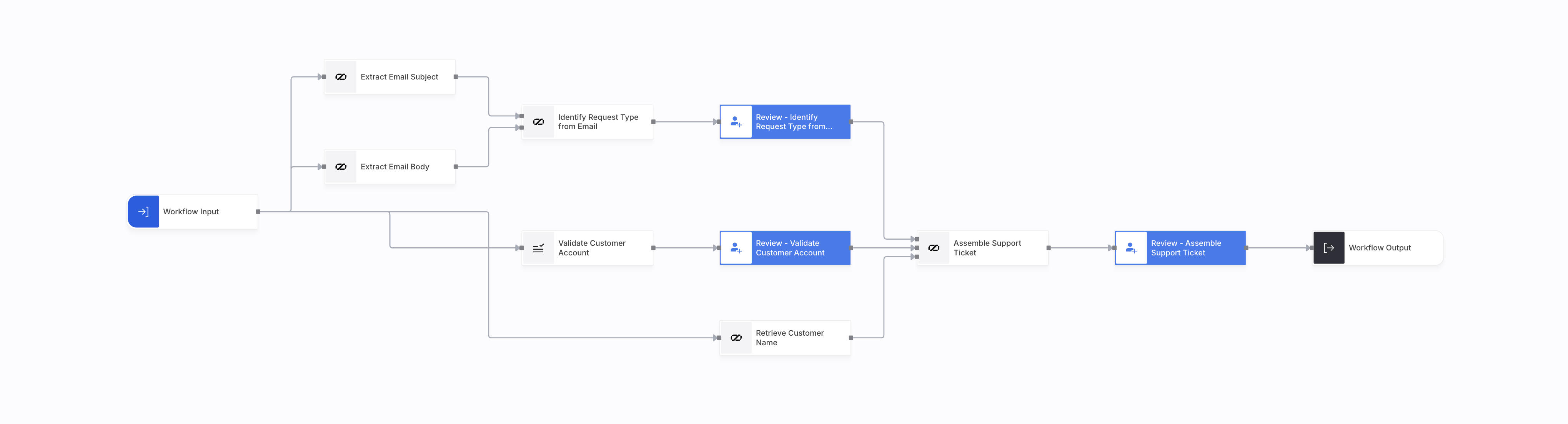}
    \caption{Workflow 1 - Extracting and classifying email content before ticket assembly.}
    \label{fig:cs1v1}
\end{figure}

Candidate $W_1$ is shown in Figure~\ref{fig:cs1v1}. The Workflow Tasks are:
\begin{itemize}[label={}]
    \item \textit{Extract Email Subject}: $p=0.95$, $d=1500+800$, $\text{cost}=1.6 \times 10^{-4}$
    \item \textit{Extract Email Body}: $d=50+800$
    \item \textit{Identify Request Type from Email}: $p=0.9$, $d=1500$, $\text{cost}=1.6 \times 10^{-4}$
    \item \textit{Validate Customer Account}: $d=200+800$
    \item \textit{Review - Identify Request Type from Email}: $p=0.95$, $q=0.7$, $d=1500$, $\text{cost}=8.0 \times 10^{-4}$
    \item \textit{Review - Validate Customer Account}: $p=0.99$, $q=0.7$, $d=500$, $\text{cost}=7.0 \times 10^{-4}$
    \item \textit{Retrieve Customer Name}: $d=200+800$
    \item \textit{Assemble Support Ticket}: $d=10$
    \item \textit{Review - Assemble Support Ticket}: $p=0.95$, $q=0.7$, $d=800$, $\text{cost}=7.0 \times 10^{-4}$
\end{itemize}

\subsection{Workflow 2}
\begin{figure}[H] 
    \centering
    \includegraphics[width=1\textwidth]{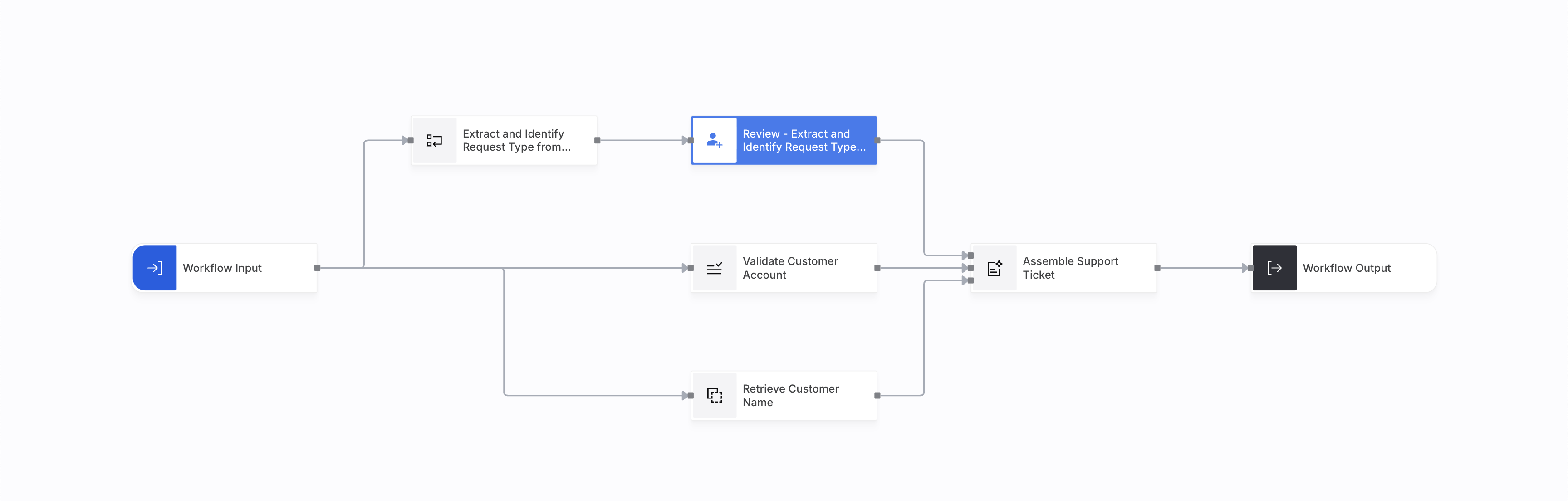}
    \caption{Workflow 2 - Compact design with fewer reviews.}
    \label{fig:cs1v2}
\end{figure}

Candidate $W_2$ is shown in Figure~\ref{fig:cs1v2}. The Workflow Tasks are:
\begin{itemize}[label={}]
    \item \textit{Extract and Identify Request Type from Email}: $p=0.9$, $d=1500+800$, $\text{cost}=1.6 \times 10^{-4}$
    \item \textit{Review - Extract and Identify Request Type from Email}: $p=0.95$, $q=0.7$, $d=1500$, $\text{cost}=8.0 \times 10^{-4}$
    \item \textit{Validate Customer Account}: $d=200+800$
    \item \textit{Retrieve Customer Name}: $d=200+800$
    \item \textit{Assemble Support Ticket}: $d=10$
\end{itemize}

\subsection{Workflow 3}
\begin{figure}[H] 
    \centering
    \includegraphics[width=1\textwidth]{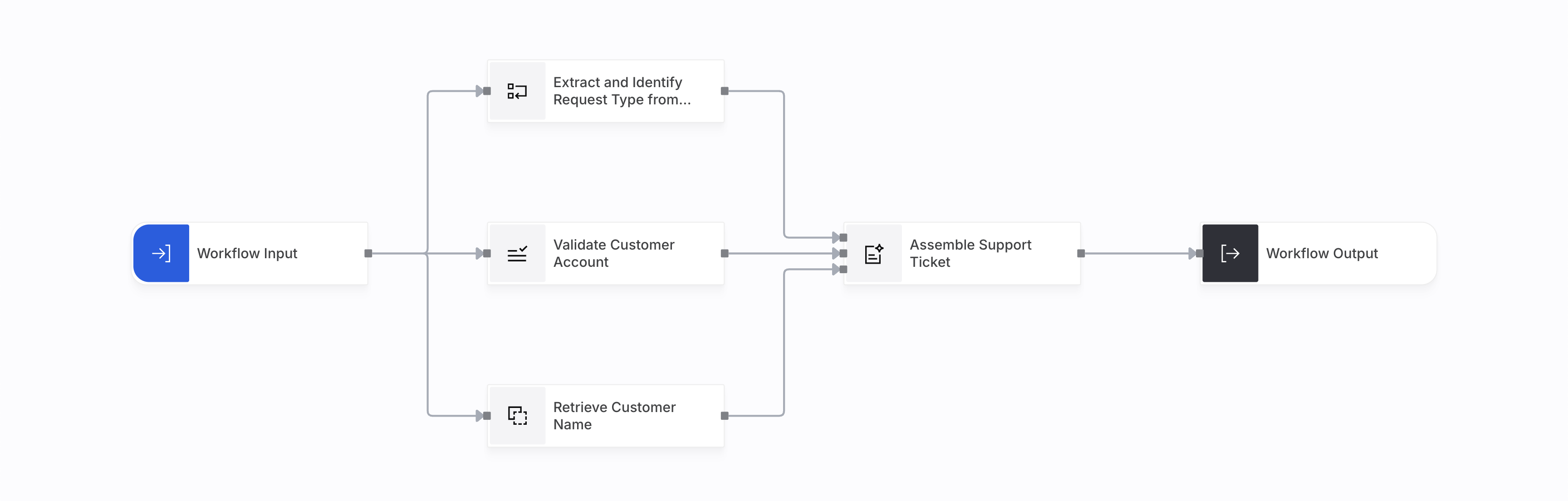}
    \caption{Workflow 3 - Merged review.}
    \label{fig:cs1v3}
\end{figure}

Candidate $W_3$ is shown in Figure~\ref{fig:cs1v3}. The Workflow Tasks are:
\begin{itemize}[label={}]
    \item \textit{Extract and Identify Request Type from Email}: $p=0.9 \times 0.95 + (1-0.9) \times 0.7 = 0.925$, $d=1500+1500+800$, $\text{cost}=9.6 \times 10^{-4}$
    \item \textit{Validate Customer Account}: $d=200+800$
    \item \textit{Retrieve Customer Name}: $d=200+800$
    \item \textit{Assemble Support Ticket}: $d=10$
\end{itemize}

This Workflow merges the nodes \textit{Extract and Identify Request Type from Email} and \textit{Review of Extract and Identify Request Type from Email}.

\subsection{Analysis}
\begin{table}[H]
\centering
\begin{tabular}{lccc}
\toprule
\textbf{Benchmark} & \textbf{Workflow 1} & \textbf{Workflow 2} & \textbf{Workflow 3} \\
\midrule
Cost (\$) & $2.52 \times 10^{-3}$ & $\boldsymbol{9.6 \times 10^{-4}}$ & $\boldsymbol{9.6 \times 10^{-4}}$\\
Max duration (ms) & 6110.0 & \textbf{3810.0} & \textbf{3810.0} \\
Success probability & \textbf{0.9262} & 0.9250 & 0.9250 \\
Reward $\mathcal{R}$ (\$) & 0.8495 & \textbf{0.8500} & \textbf{0.8500} \\
CIP & 0.2525 & \textbf{0.2520} & 0.2570 \\
SIP & \textbf{0.2500} & 0.2573 & 0.2848 \\
Penalty $\mathcal{L}$ & \textbf{0.2513} & 0.2547 & 0.2713 \\
\bottomrule
\end{tabular}
\caption{Benchmark evaluation for the three candidate Workflows.}
\label{tab:cs1_benchmark}
\end{table}

Table~\ref{tab:cs1_benchmark} summarizes the metrics. The expected Reward values $\mathcal{R}(W_1) = \$0.8495$ and $\mathcal{R}(W_2) = \$0.8500$ establish that $W_2 \succ W_1$. $W_2$ and $W_3$ achieve identical Reward scores. However, their topologies differ: in $W_3$, merging the review with the action prevents independent evaluation of the review’s contribution. Observability is therefore reduced (because only output information is displayed), and the Penalty increases. For these reasons, Workflow 2 emerges as the most balanced option: lower cost and shorter runtime, with high probability of success preserved. Finally, note that our model prioritizes shorter execution times over higher reliability. If reliability were valued more highly, no corrective term would be required in the equations: adjusting the gain parameters, especially the CLV, would naturally shift the optimization. While serverless costs dominate in low-value B2C cases, for high-value enterprise scenarios, reliability becomes paramount.

\section{Conclusion}
In this paper, we introduced the Opus Workflow Evaluation Framework, a unified probabilistic-normative system for measuring and optimizing the quality of AI-driven Workflows. The framework formalizes the Opus Workflow Reward, a probabilistic expectation over success, cost, and gain, and the Opus Workflow Normative Penalties, a continuous set of structural quality measures grounded in Cohesion, Coupling, Observability, and Information Hygiene. Together, these components transform Workflow assessment into a quantitative, optimization-oriented process.

\vspace{1em}

Empirical evaluations confirm that the framework is directly applicable to live Workflow builder systems, where it enables the consistent evaluation, ranking, and refinement of Workflows in production environments. By quantifying performance and structure within a unified formulation, the framework supports objective comparison and optimization across heterogeneous processes. Future work will extend this foundation toward Reinforcement Learning settings, where the Opus Reward and Normative Penalties will serve as feedback signals guiding autonomous Workflow discovery and continuous improvement. This progression will position the Opus Workflow Evaluation Framework as a cornerstone for self-optimizing Workflow automation systems.

\newpage

\appendix

\section{Appendix}

\subsection{Conditional Workflows}

Our current framework does not natively support conditional branching within Workflows. This design decision was made to preserve the simplicity and tractability of the overall system. However, conditional branching is ubiquitous in real-world scenarios, and therefore it is essential to provide a principled way to incorporate such behavior into our model.

\vspace{1em}

Let us assume that a conditional Workflow \( W \) can give rise to a finite set of deterministic execution scenarios \( \{W_1, W_2, \dots, W_n\} \), depending on the evaluation of internal conditions. Each \( W_i \) represents a fully resolved, deterministic Workflow that may be followed in practice.

\vspace{1em}

To account for the uncertainty inherent in the conditional branching, we associate to each scenario \( W_i \) a probability \( p_i \in [0,1] \), such that \( \sum_{i=1}^n p_i = 1 \). These probabilities reflect the likelihood of each branch being taken, and may either be known a priori, learned from data, or inferred from domain knowledge.

\vspace{1em}

Given a Reward function \( \mathcal{R}(W) \) that evaluates deterministic Workflows, we define the Reward of the conditional Workflow \( W \) as the expected Reward over all its possible realizations:

\begin{equation}
\mathcal{R}(W) = \sum_{i=1}^n p_i \cdot \mathcal{R}(W_i)
\end{equation}

Since our original Reward function \( \mathcal{R} \) is itself expectation-based, this extension provides a natural and elegant generalization of the framework. Moreover, this formulation is easy to implement in practice, and seamlessly integrates with the existing system.

\subsection{Computation of Releasable Resource Peaks}

\paragraph{ASAP start-times and finish-times (recurrence).}
Under an ASAP schedule the start time \(s_v\) and finish time \(f_v\) of each Task \(v\) are defined recursively by
\begin{align}
    s_v &=
    \begin{cases}
      0 & \text{if }\mathrm{pred}(v)=\varnothing\\[4pt]
      \displaystyle \max_{u\in\mathrm{pred}(v)} f_u & \text{otherwise}
    \end{cases}
    \label{eq:start-rec}\\[6pt]
    f_v &= s_v + d_v
    \label{eq:finish-def}
\end{align}

Since \(W\) is a DAG and all \(d_v\) are finite, the recurrence~\eqref{eq:start-rec} defines \(s_v\) and \(f_v\) uniquely for every \(v\).

\vspace{1em}

For any time \(t\in\mathbb{R}^+\) we take the standard half-open convention for activity intervals and define
\begin{equation}
    V_t \;=\; \{\, v\in V_{\text{task}} \;:\; s_v \le t < f_v \,\}
    \label{eq:Vt-def}
\end{equation}
Equivalently, a Task \(v\) is active at \(t\) if and only if \(t\in [s_v,f_v)\).

\vspace{1em}

By defining a min-heap \(\mathcal{H}\) that stores finish events of the form \((f_v, v, r_v^{(r)})\) (sorted by finish time \(f_v\)), we can compute \(R^{(r)}(W)\) using an event-driven algorithm.  At each step, \(\mathtt{R}_{\mathrm{curr}}\) stores the current sum of active releasable resources, and \(\mathtt{R}_{\mathrm{max}}\) stores the componentwise maximum observed so far.

\begin{algorithm}[H]
\caption{Computation of $R^{(r)}(W)$ under an ASAP schedule}
\begin{algorithmic}[1]
\State Initialize $\mathtt{pred\_count}[v] \gets |\mathrm{pred}(v)|$ for all $v \in V$
\State $\mathtt{R}_{\mathrm{curr}} \gets 0$, $\mathtt{R}_{\mathrm{max}} \gets 0$, $\mathcal{H} \gets \emptyset$
\ForAll{$v$ with $\mathtt{pred\_count}[v] = 0$} 
    \State $\mathtt{R}_{\mathrm{curr}} \mathrel{+}= r_v^{(r)}$
    \State Push $(f_v = d_v, v, r_v^{(r)})$ into $\mathcal{H}$
\EndFor
\State $\mathtt{R}_{\mathrm{max}} \gets \mathtt{R}_{\mathrm{curr}}$
\While{$\mathcal{H}$ not empty}
    \State $(t, v, r_v^{(r)}) \gets$ Pop earliest event from $\mathcal{H}$
    \State $\mathtt{R}_{\mathrm{curr}} \mathrel{-}= r_v^{(r)}$
    \ForAll{$w \in \mathrm{succ}(v)$}
        \State $\mathtt{pred\_count}[w] \mathrel{-}= 1$
        \If{$\mathtt{pred\_count}[w] = 0$}
            \State $\mathtt{R}_{\mathrm{curr}} \mathrel{+}= r_w^{(r)}$
            \State Push $(t + d_w, w, r_w^{(r)})$ into $\mathcal{H}$ \Comment{$f_w = s_w + d_w$ with $s_w = t$}
        \EndIf
    \EndFor
    \State $\mathtt{R}_{\mathrm{max}} \gets \max(\mathtt{R}_{\mathrm{max}}, \mathtt{R}_{\mathrm{curr}})$
\EndWhile
\State \Return $\mathtt{R}_{\mathrm{max}}$
\end{algorithmic}
\end{algorithm}

\subsection{Single Responsibility Principle (SRP): defining a contextual level of atomicity}

The conceptual decomposition of Tasks, ranging from high-level responsibilities (e.g. RACI roles) to atomic units of work (e.g. field-level transactions), is inspired by frameworks used in Business Process Management literature. It outlines four conceptual levels of Task decomposition, ranging from high-level roles (Level 1) to atomic actions (Level 4). Within this framework, we present our notion of singular responsibility not as a fixed level of atomicity, but rather as a contextual one. Depending on the Workflow, the actor, and the system’s perspective, the optimal level of decomposition may vary.

\vspace{1em}

For instance, what qualifies as a single-responsibility Task for an AI system (such as “Send email”) may fall under Level 2 (Task level), whereas a software engineer might model this same Task at Level 4, breaking it down into protocol-level operations (e.g. TCP handshake, SMTP commands). Thus, single responsibility is not an absolute metric; it is relative to the granularity that makes sense for the agent performing the Task.

\vspace{1em}

This aligns with Weske’s interpretation of atomicity \cite{weske_bpm}: “An activity is atomic if it cannot be sensibly subdivided given the process context and stakeholder objectives.” Hence, the goal is not to find the “true atoms” of work universally, but to identify the optimal level of decomposition for a given purpose and context: striking a balance between meaningful action and implementation detail.

\vspace{1em}

This framework aligns with established BPM literature, where hierarchical Task decomposition is well-documented. Scheer's ARIS \cite{scheer_1999} methodology employs similar multi-level decomposition from organizational roles down to atomic business Tasks, while Van der Aalst \cite{aalst_process_mining} describes comparable layered approaches in process mining, distinguishing between high-level business processes and their constituent atomic activities.

\paragraph{Discrete formalization}

The Single Responsibility Principle can be formalized as follows.

\vspace{1em}

We define the granularity assignment function as:

\begin{equation}
\psi : V \times \mathcal{W} \to L = \{L_1, L_2, L_3, L_4\}
\end{equation}

where $V$ is the ensemble of all valid vertices and $\mathcal{W}$ the ensemble of all valid Workflows, and $L$ the set of four classic levels of granularity. 

\vspace{1em}

We define $L^*_W$ as the optimal level of granularity of a Workflow, stating that:
\begin{itemize}[label={}]
    \item A given Task $v \in V$ in the context $W$ has one clear responsibility at level $L^*_W$;
    
    \item Its subdivision to a deeper level than $L^*_W$  would add confusion and not clarity;
    
    \item Task $v$ is complete (accomplishes its purpose) at level $L^*_W$. 
\end{itemize}

For a given $W$ in $\mathcal{W}$, SRP is valid if and only if:
\begin{equation}
\psi(v,W) = L^*_W,
\quad \forall \; v \in V
\end{equation}
\paragraph{Continuous formalization and intuition}
While the discrete definition provides structural clarity, it does not capture the continuum of Task granularity observed in real systems. To address this, we introduce a continuous version of the granularity function:

\begin{equation}
\psi_c : V \times \mathcal{W} \to [0, 1]
\end{equation}

where $\psi_c(v,W)$ represents the degree of atomicity of a given Task $v$ within Workflow $W$. 

\vspace{1em}

We define $\lambda^*_W$ as the optimal continuous granularity target for Workflow $W$, such that:

\begin{equation}
\psi_c(v,W) = \lambda^*_W
\quad \forall \; v \in V
\end{equation}

Intuitively:
\begin{itemize}[label={}]
    \item $\lambda^*_W \approx 1$ implies that nodes should be treated as atomic operations (maximal decomposition);
    \item $\lambda^*_W \approx 0$ implies that nodes should be merged or abstracted into a higher-level responsibility;
    \item Intermediate values express contextual balance between decomposition depth and cohesion.
\end{itemize}

\newpage

This continuous mapping allows SRP to interact more easily with optimization-based frameworks.

It is straightforward to observe from Eq.~\ref{eq:CIP} that, in order to minimize the CIP, each node’s Coupling $\mathrm{Cp}(v)$ should approach the value of $\alpha_{\mathrm{Ch}}$. Since $\mathrm{Cp}(v)$ can be interpreted as an inverse indicator of atomicity (high Coupling implying lower independence) $\alpha_{\mathrm{Ch}}$ can therefore be regarded as the target degree of atomicity of the Workflow.

\vspace{1em}

Formally, we may identify:
\begin{equation}
\lambda^*_W =\alpha_{\text{Ch}}
\end{equation}

Hence, the system’s structural balance between cohesion and Coupling directly defines its semantic target level of decomposition.
    
\subsection{Workflow Composition}

We define two binary operations on Workflows: parallel composition, denoted by \( W_1 \parallel W_2 \), and sequential composition, denoted by \( W_1 \circ W_2 \). Each operation produces a new Workflow \( W = (G, \Phi) \), preserving the acyclic graph structure and ensuring semantic consistency of Task dependencies.

\vspace{1em}

Let \( W_1 = (G_1, \Phi_1) \), \( W_2 = (G_2, \Phi_2) \), with \( G_i = (V_i, E_i) \) for \( i = 1, 2 \). We assume that the node sets \( V_1 \) and \( V_2 \) are disjoint (if not, they are renamed via canonical relabeling).

\paragraph{Parallel Composition}

The parallel composition corresponds to the concurrent execution of \( W_1 \) and \( W_2 \), without introducing additional control flow between them.

\[
W_1 \parallel W_2 = (G, \Phi), \quad \text{where}
\]
\[
G = (V_1 \cup V_2, \; E_1 \cup E_2), \quad \Phi = \Phi_1 \cup \Phi_2
\]

This operation preserves the internal structure of both Workflows. Workflow Input, Task, and Output nodes are combined:
\[
V_{\text{in}} = V_{\text{in}}^{(1)} \cup V_{\text{in}}^{(2)}, \quad V_{\text{task}} = V_{\text{task}}^{(1)} \cup V_{\text{task}}^{(2)}, \quad V_{\text{out}} = V_{\text{out}}^{(1)} \cup V_{\text{out}}^{(2)}
\]

The resulting Workflow executes both components independently. No edges are added between \( V_1 \) and \( V_2 \).

\paragraph{Sequential Composition}

The sequential composition connects two Workflows \( W_1 = (G_1, \Phi_1) \) and \( W_2 = (G_2, \Phi_2) \), such that the outputs of \( W_1 \) feed directly into the inputs of \( W_2 \), bypassing any explicit intermediary nodes. This composition is only defined when the following compatibility condition holds:
\[
V^{(2)}_{\text{in}} \subseteq V^{(1)}_{\text{out}}
\]

\(V^{(2)}_{\text{in}} \cap V^{(1)}_{\text{out}} = V^{(2)}_{\text{in}} \) is the set of interface nodes to be removed. These nodes are entirely eliminated from the composed Workflow, and all connections to or from them are rewired.

\newpage

The composed Workflow is:
\[
W_1 \circ W_2 = (G, \Phi), \quad \text{with } G = (V, E), \quad \Phi = \Phi_1 \cup \Phi_2
\]

where:
\[
V = (V_1 \cup V_2) \setminus V^{(2)}_{\text{in}}
\]

The edge set \( E \) is defined as:
\[
E = (E_1 \cup E_2) \setminus \left( E_{\text{in}} \cup E_{\text{out}} \right) \cup E_{\text{bridge}}
\]

with:
\begin{align*}
E_{\text{in}} &= \{ (u \to v) \in E_1 \mid v \in V^{(2)}_{\text{in}} \} \\
E_{\text{out}} &= \{ (v \to w) \in E_2 \mid v \in V^{(2)}_{\text{in}} \} \\
E_{\text{bridge}} &= \{ (u \to w) \mid (u \to v) \in E_1, \; (v \to w) \in E_2, \; v \in V^{(2)}_{\text{in}} \}
\end{align*}

In other words:
\begin{itemize}[label={}]
  \item All nodes in \( V^{(2)}_{\text{in}} \) are removed from the graph.
  \item All edges pointing to or from these nodes are removed.
  \item For each pair of edges \( (u \to v) \in E_1 \), \( (v \to w) \in E_2 \), where \( v \in V^{(2)}_{\text{in}} \), we introduce a new edge \( (u \to w) \in E_{\text{bridge}} \), thereby directly connecting predecessors of \( v \) in \( W_1 \) to its successors in \( W_2 \).
\end{itemize}

This rewiring ensures that the data flow is preserved while eliminating redundant intermediate nodes.

\vspace{1em}

This operation strictly preserves the DAG property of the resulting Workflow, assuming both \( G_1 \) and \( G_2 \) are DAGs and that the rewiring does not introduce backward edges.

\paragraph{Remarks}
\begin{itemize}[label={}]
  \item Sequential composition is not commutative: $\exists \; W_1, W_2 \quad \text{such that} \quad  W_1 \circ W_2 \neq W_2 \circ W_1$.
  \item This composition preserves acyclicity, assuming that both \( G_1 \) and \( G_2 \) are DAGs and that the bridge edges respect temporal ordering.
\end{itemize}

\newpage
\subsection{Cost Properties}
We define two levels of Workflow equivalence:

\begin{itemize}[label={}]
  \item \textbf{Weak equivalence}, denoted \( W_1 \sim W_2 \), holds if two Workflows share the same Input and Output interfaces. Their internal procedures (Tasks) or execution behavior may differ.

  \item \textbf{Strong equivalence}, denoted \( W_1 \equiv W_2 \), holds if two Workflows are weakly equivalent and exhibit identical probabilistic behavior when fully decomposed into atomic actions. They are indistinguishable in terms of both structure and outcome distribution at the most granular level of execution.
\end{itemize}

Let \( \mathcal{W} \) denote the set of all valid Workflows, and let \( C : \mathcal{W} \to \mathbb{R} \) be a cost metric that assigns a real number to each Workflow.

\paragraph{Non-triviality}  
Not all Workflows should have the same cost; at least two Workflows must differ in their cost values.  
\[
\exists \;W_1, W_2 \in \mathcal{W} \quad \text{such that} \quad C(W_1) \neq C(W_2)
\]

\paragraph{Implementation Sensitivity}  
Two weakly equivalent Workflows can still have different costs due to differences in structure or implementation.  
\[
\exists \; W_1, W_2 \in \mathcal{W} \quad \text{such that} \quad W_1 \sim W_2 \quad \text{and} \quad C(W_1) \neq C(W_2)
\]

\paragraph{Cost Invariance}  
Two strongly equivalent Workflows must have identical costs.  
\[
\forall \;W_1, W_2 \in \mathcal{W}, \quad \quad W_1 \equiv W_2 \;\Rightarrow\; C(W_1) = C(W_2)
\]

\paragraph{Sub-additivity}  
Combining two Workflows may share resources or eliminate redundancies, resulting in a total cost that is lower than the sum of their individual costs.
\[
\forall \; * \in \{\circ, ||\}, \quad \forall \; W_1, W_2 \in \mathcal{W}, \quad \quad C(W_1 * W_2) \leq C(W_1) + C(W_2)
\]

\paragraph{Context Sensitivity}  
Appending the same Workflow to two Workflows with equal cost may yield different overall costs, depending on the structure of the prefix.  
\[
\forall \; * \in \{\circ, ||\}, \quad \exists \; W, V_1, V_2 \in \mathcal{W} \quad \text{such that} \quad C(V_1) = C(V_2) \quad \text{and} \quad C(V_1*W) \neq C(V_2*W)
\]

\paragraph{Order Sensitivity}  
The order in which sub-workflows are composed affects the total cost; reordering operations can increase or reduce cost.  
\[
\exists \; W_1, W_2 \in \mathcal{W} \quad \text{such that} \quad C(W_1 \circ W_2) \neq C(W_2 \circ W_1)
\]

\paragraph{Parallel Commutativity}  
There is no order in parallelizing sub-workflows,  hence parallelizing in one way or another does not affect the total cost.
 \[
\forall \; W_1, W_2 \in \mathcal{W}, \quad W_1 || W_2 \equiv W_2 || W_1
\]
By Cost Invariance for strongly equivalent Workflows, it follows that:
\[
\forall \; W_1, W_2 \in \mathcal{W}, \quad C(W_1 || W_2) = C(W_2 || W_1)
\]

\end{document}